\documentclass[10pt]{article}
\usepackage[margin=1in]{geometry}

\usepackage[affil-it]{authblk} 
\usepackage{etoolbox}
\usepackage{lmodern}
\usepackage{rotating}
\makeatletter
\patchcmd{\@maketitle}{\LARGE \@title}{\fontsize{14}{17.2}\selectfont\@title}{}{}
\makeatother

\usepackage{sectsty}

\sectionfont{\fontsize{13}{15}\selectfont}
\subsectionfont{\fontsize{12}{15}\selectfont}

\usepackage[font=footnotesize]{caption}
\usepackage{authblk}
\usepackage{microtype}
\usepackage{graphicx}
\usepackage{subfigure}
\usepackage{booktabs} 
\usepackage{amsmath,amssymb, amsthm, bm}
\usepackage{color,soul}

\usepackage{setspace}
\usepackage{gensymb}
\usepackage{bbm}
\usepackage[colorlinks=true,allcolors=blue]{hyperref}
\usepackage{natbib}
\usepackage{url}
\usepackage{wrapfig}

\usepackage{mathtools}

\usepackage[ruled,vlined]{algorithm2e}
\usepackage{amsmath}
\usepackage[makeroom]{cancel}

\usepackage{makecell}
\DeclarePairedDelimiterX{\infdivx}[2]{(}{)}{%
  #1\;\delimsize\|\;#2%
}

\newcommand{\norm}[1]{\left\lVert#1\right\rVert}

\newcommand{\btheta}{\boldsymbol{\theta}}

\newtheorem{theorem}{Theorem}
\newtheorem{definition}[theorem]{Definition}

\newtheorem{remark}[theorem]{Remark}

\newcommand{\bz}{\mathbf{z}}
\newcommand{\bI}{\mathbf{I}}

\newcommand{\hSk}{\widehat{S}_k}

\newcommand{\ball}{{\rm I\!B}}

\title{\textbf{SALR: Sharpness-aware Learning Rate Scheduler for Improved Generalization}}
\date{\vspace{-7ex}}

\author[1]{Xubo Yue}
\author[2]{Maher Nouiehed}
\author[1]{Raed Al Kontar}

\affil[1]{Industrial and Operations Engineering\\University of Michigan, Ann Arbor}
\affil[2]{American University of Beirut, Lebanon}

\begin{document}
\maketitle

\begin{abstract}
In an effort to improve generalization in deep learning and automate the process of learning rate scheduling, we propose SALR: a sharpness-aware learning rate update technique designed to recover flat minimizers. Our method dynamically updates the learning rate of gradient-based optimizers based on the local sharpness of the loss function. This allows optimizers to automatically increase learning rates at sharp valleys to increase the chance of escaping them. We demonstrate the effectiveness of SALR when adopted by various algorithms over a broad range of networks. Our experiments indicate that SALR improves generalization, converges faster, and drives solutions to significantly flatter regions. 
\end{abstract}

\section{Introduction}
Generalization in deep learning has recently been an active area of research. The efforts to improve generalization over the past two decades have brought upon many cornerstone advances and techniques; be it dropout \citep{gal2016dropout}, batch-normalization \citep{ioffe2015batch}, data-augmentation \citep{shorten2019survey}, weight decay \citep{loshchilov2019decoupled}, adaptive gradient-based optimization \citep{kingma2014adam}, architecture design and search \citep{radosavovic2020designing}, ensembles and their Bayesian counterparts \citep{garipov2018loss, izmailov2018averaging}, amongst many others. Yet, recently, researchers have discovered that the concept of sharpness/flatness plays a fundamental role in generalization.

Though sharpness was first discussed in the context of neural networks in the early work of \citep{hochreiter1997flat}, it was only brought to the forefront of deep learning research after the seminal paper by \citep{keskar2016large}. While trying to investigate decreased generalization performance when large batch sizes are used in stochastic gradient descent (SGD) \citep{lecun2012efficient}, \citep{keskar2016large} noticed that this phenomena can be justified by the ability  of smaller batches to reach flat minimizers. Such flat minimizers in turn, generalize well as they are robust to low precision arithmetic or noise in the parameter space \citep{dinh2017sharp, kleinberg2018alternative}; 


Since then, the generalization ability of flat minimizers has been repeatedly observed in many recent works \citep{neyshabur2017exploring, goyal2017accurate, li2018visualizing, izmailov2018averaging, chaudhari2019entropy, foret2021sharpness}. Indeed, flat minimizers can potentially tie together many of the aforementioned approaches aimed at generalization. For instance, (1) higher gradient variance, when batches are small increases the probability to avoid sharp regions (same can be said for SGD compared to GD) \citep{kleinberg2018alternative} (2) averaging over multiple hypotheses leads to wider optima in ensembles and Bayesian deep learning \citep{izmailov2018averaging} (3) regularization techniques such as dropout and over-parameterization can adjust the loss landscape into one that allows first order methods to favor wide valleys \citep{chaudhari2019entropy, allen2019convergence}.

In this paper we study the problem of developing an algorithm that can converge to flat minimizers. Specifically, we introduce SALR: a sharpness aware learning rate designed to explore the loss-surface of an objective function and avoid undesired sharp local minima. SALR dynamically updates the learning rate based on the sharpness of the neighborhood of the current solution.
The idea is simple: automatically increase the learning rates at relatively sharp valleys in an effort to escape them. A key features of SALR is its ability to be adopted by any gradient based method such as \textbf{Adagrad} \citep{duchi2011adaptive}, \textbf{Adam} \citep{kingma2014adam} \textbf{and also into recent approaches towards escaping sharp valleys such as Entropy-SGD} \citep{chaudhari2019entropy}. Our contributions are summarized below:

\begin{itemize}
    \item Motivated by recent results on the improved generalization capability of flat minimizers, we propose SALR: a dynamic learning rate update mechanism that utilizes the sharpness of the underlying landscape. In particular, our mechanism increases the learning rate in sharp regions to increase the chance of escaping them. Our framework can be adopted by a wide variety of gradient-based method. We then show that GD-SALR (gradient descent adopting our learning rate schedule) can escape strongly convex local minimizers.
    \item We demonstrate the improved generalization achieved when adopting SALR on various optimization methods (SGD, Adam, Entropy-SGD) applied to a wide range of applications (Image classifications, Text prediction, Fine-tuning) and using a variety of network structures and datasets.
    \item \textbf{SALR circumvents one of the key practical challenges in deep learning: setting a heuristic learning rate schedule}. Instead, SALR dynamically chooses learning rates to recover flat solutions. 
\end{itemize}

\subsection{Related Work}
\label{subsec:literature}
From a theoretical perspective, generalization of deep learning solutions has been explained through multiple lenses. One of which is uniform stability \citep{bottou2005line, bottou2008tradeoffs, hardt2016train, gonen2017fast, bottou2018optimization}. An algorithm is uniformly stable if for all data sets differing in only one element, nearly the same outputs will be produced \citep{bousquet2002stability}. \citep{hardt2016train} show that SGD satisfies this property and derive a generalization bound for models learned with SGD. From a different viewpoint, \citep{choromanska2015loss, kawaguchi2016deep, poggio2017and, mohri2018foundations} attribute generalization to the complexity of the hypothesis-space. 
Using measures like Rademacher complexity \citep{mohri2009rademacher} and the Vapnik-Chervonenkis (VC) dimension \citep{sontag1998vc}, the former works
show that deep hypothesis spaces are typically more advantageous in representing complex functions. Besides that, the importance of flatness on generalization has been theoretically highlighted through PAC-Bayes bounds \citep{dziugaite2017computing, neyshabur2017pac,wang2018identifying}. These papers highlight the ability to derive non-vacuous
generalization bounds based on the sharpness of a model class while arguing that relatively flat solutions yield tight bounds. 


From an algorithmic perspective, approaches to recover flat minima are still limited. Most notably, \citep{chaudhari2019entropy} developed the Entropy-SGD algorithm. Entropy-SGD defines a local-entropy-based objective which smoothens the energy landscape based on its local geometry. This in turn allows SGD to attain flatter solutions. Indeed, this approach was motivated by earlier work in statistical physics \citep{baldassi2015subdominant, baldassi2016unreasonable} which proves the existence of non-isolated solutions that generalize well in networks with discrete weights. Such non-isolated solutions correspond to flat minima in continuous settings. The authors then propose a set of approaches based on ensembles and replicas of the loss to favor wide solutions. Not too far, recent methods in Bayesian deep learning (BDL) have also shown potential to recover flat minima. BDL basically averages over multiple hypotheses weighted by their posterior probabilities (ensembles being a special case of BDL \citep{izmailov2018averaging}). One example, is the stochastic weighted averaging (SWA) algorithm proposed by \citep{izmailov2018averaging}. SWA simply averages over multiple points along the trajectory of SGD to potentially find flatter solutions compared to SGD. Another example is the SWA-Gaussian (SWAG). SWAG defines a Gaussian posterior approximation over neural network weights. Afterwards, samples are taken from the approximated distribution to perform Bayesian model averaging \citep{maddox2019simple}. Besides Entropy-SGD and SWA, \citep{wen2018smoothout} propose a method SmoothOut to smooth out sharp minima by averaging over multiple perturbed copies of the landscape. More recently, \citep{foret2021sharpness} proposed a Sharpness-Aware Minimization (SAM) method for finding flat minimizer. In particular, SAM solves a min-max optimization problem that minimizes the maximum loss when parameters are allowed a small perturbation. 


Another recent work that motivates our framework is the method proposed by~\citep{patel2017impact}. Based on the aformentioned observations in  \citep{keskar2016large}, \citep{patel2017impact} shows that the learning rate lower-bound threshold for the divergence of batch SGD, run on quadratic optimization problems, increases for larger batch-sizes. More specifically, in general non-convex settings, given a problem with $N$ local minimizers, one can compute $N$ lower bound thresholds for local divergence of batch SGD. The number of minimizers for which batch SGD can converge is non-decreasing in the batch size. This is used to explain the tendency of low-batch SGD to converge to flatter minimizers compared to large-batch SGD. The former result links the choice of batch size and its effect on generalization to the choice of the learning rate. With the latter being a tunable parameter, to our knowledge, developing a dynamic choice of the learning rate that targets convergence to flat minimizers has not been studied.

\section{General Framework} \label{sec:General Framework}
In this paper, we propose a framework that dynamically chooses a {\it Sharpness-Aware Learning Rate} to promote convergence to flat minimizers. More specifically, our proposed method locally approximates sharpness at the current iterate and dynamically adjusts the learning rate accordingly. In sharp regions, relatively large learning rates are attained to increase the chance of escaping that region. In contrast, when the current iterate belongs to a flat region, our method returns a relatively small learning rate to guarantee convergence. Our framework can be adopted by any local search descent method and is detailed in Algorithm 1.

\begin{algorithm}[htb]
    \caption{Sharpness-Aware Learning Rate (SALR) Framework}
\SetAlgoLined
	\KwData{Starting point $\btheta_0$, initial learning rate $\eta_0$, number of iterations $K$.}
    \For{$k=0, 1, \ldots, K$}{
         Estimate $\hSk$, the local sharpness around the current iterate $\btheta_k$\;
         Set $\eta_k= \eta_0 \dfrac{\hSk}{\mbox{Median} \left\{\widehat{S}_i\right\}_{i=1}^k}$\;
         Compute $\btheta_{k+1}$ using some local search descent method (Gradient Descent, Stochastic Gradient Descent, Adam, $\ldots$)\;
    }
	Return $\btheta_K$\;
	    \label{algo:0}
\end{algorithm}

\begin{figure*}[!ht]
    \centering
    \centerline{\includegraphics[width=\columnwidth]{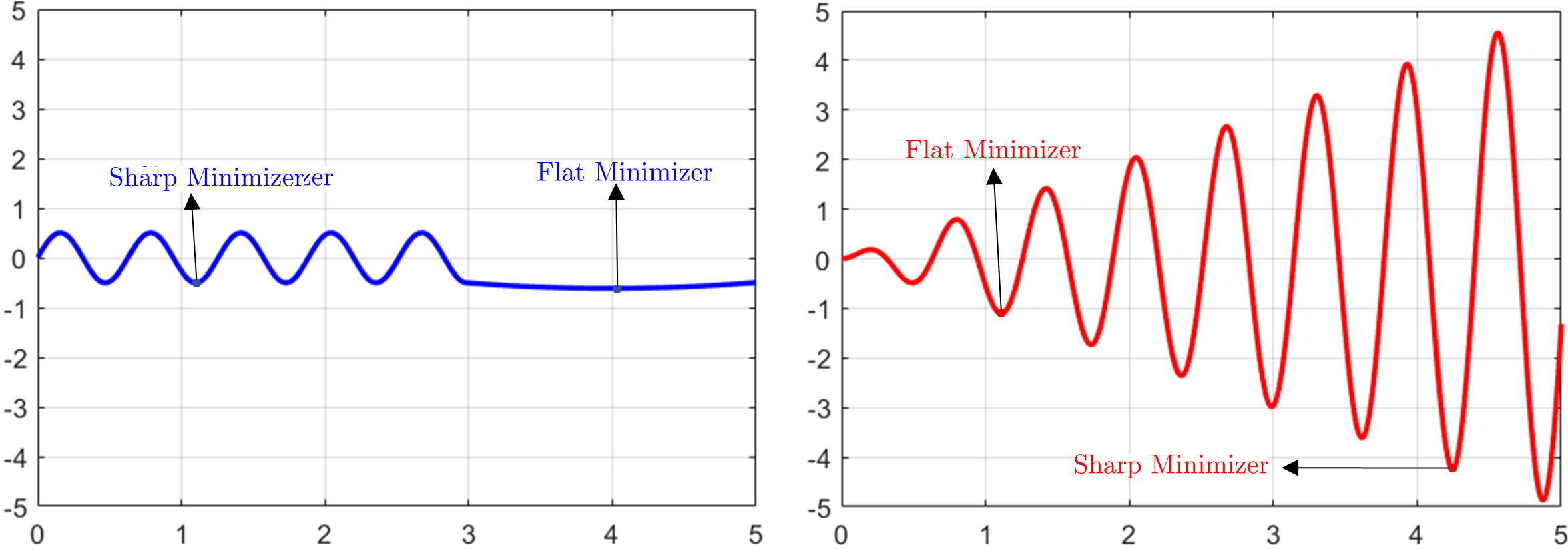}}
    \caption{Sharp/Flat minimizers relative to the landscape.}
    \label{fig:flat_sharp}
\end{figure*}

As detailed in Algorithm 1, at every iterate $k$, we compute the learning rate as a function of the local sharpness parameters $\left\{ \widehat{S}_k \right\}_{i=1}^k$. The main intuition is to have the current learning rate to be an increasing function of the current estimated sharpness. Since the scale of the sharpness at different points can vary when using different networks or datasets \citep{dinh2017sharp}, we normalize our estimated sharpness by dividing by the median of the sharpness of previous iterates. For instance in Figure~\ref{fig:flat_sharp}, despite having a similar sharpness measure, we consider the minimizer around $\btheta = 1$ to be sharp relative to the blue plot and flat relative to red plot. Normalization resolves this issue by helping our sharpness measure attain scale invariant properties.

One can think of the median of previous sharpness values as a global sharpness parameter the algorithm is trying to learn. When $k$ is sufficiently large, the variation in the global sharpness parameter among different iterates will be minimal. 
From an algorithmic perspective, SALR exploits a neighborhood around the current iterate to dynamically compute a desired learning rate while simultaneously exploring the sharpness of the landscape to refine this global sharpness parameter.

\section{Sharpness Measure} 
\label{sec:sharpness measure}
Several sharpness/flatness measures have been defined in recent literature \citep{rangamani2019scale, keskar2016large, hochreiter1997flat}. For instance, \citep{hochreiter1997flat} compute flatness by measuring the size of the connected region in the parameter space where the objective remains approximately constant. In a more recent paper, \citep{rangamani2019scale} proposed a scale invariant flatness measure based on the quotient manifold. Computing such notions for complex non-convex landscapes can be intractable in practice. In addition to the cited results, \citep{keskar2016large} quantify flatness by finding the difference between the maximum value of the loss function within a small neighborhood around a given point and the current value. More specifically, they define sharpness as follows:
\begin{align}
\label{eq:sharp_keskar}
 \phi(\varepsilon, \btheta) \triangleq \frac{S(\varepsilon, \btheta)}{1+f(\bm{\theta})}  \mbox{  and  }  S(\varepsilon, \btheta) = \max_{\bm{\theta}'\in\ball_{\varepsilon}(\bm{\theta})} f(\bm{\theta}')-f(\bm{\theta}) ,
\end{align}
where $\ball_{\varepsilon}(\bm{\theta})$ is a euclidean ball with radius $\varepsilon$ centered at $\bm{\theta}$ and $1+f(\bm{\theta})$ is a normalizing coefficient. One drawback of~\eqref{eq:sharp_keskar} is that the sharpness value around a maximizer is nearly zero. To resolve this issue, one can simply modify the sharpness measure in \eqref{eq:sharp_keskar} as follows:
\begin{align}
\label{eq:sharp_2}
S(\varepsilon, \btheta) \triangleq \max_{\bm{\theta}'\in\ball_{\varepsilon}(\bm{\theta})}f(\bm{\theta}')-\min_{\bm{\theta}'\in\ball_{\varepsilon}(\bm{\theta})}f(\bm{\theta}').
\end{align}
It can be easily shown that if $\bm{\theta}$ is a local minimizer, \eqref{eq:sharp_2} is equivalent to \eqref{eq:sharp_keskar}. Both measures defined in~\eqref{eq:sharp_keskar} and~\eqref{eq:sharp_2} require solving a possibly non-convex function which is in general NP-Hard. For computational feasibility, we provide a sharpness approximation by running $n_1$ gradient ascent and $n_2$ gradient descent steps. The resulting solutions are used to approximate the maximization and minimization optimization problems. Here we note that our definition for sharpness does not include a normalizing coefficient, as $\mbox{median} \left\{\widehat{S}_i\right\}_{i=1}^k$ in Algorithm \ref{algo:0} plays this role. The details of the approximation are shown in the Definition \ref{def:sharp_approx}.


\begin{definition}\label{def:sharp_approx}
Given $\btheta \in \mathbb{R}^n$, iteration numbers $n_1$ and $n_2$, and step-size $\gamma$, we define the sharpness measure
\begin{align*}
    \widehat{S}(\btheta) &\triangleq  f(\bm{\theta}^{(n_2)}_{k,+}) - f(\bm{\theta}_k) + f(\bm{\theta}_k)  - f(\bm{\theta}^{(n_1)}_{k,-})\\
    &=f(\bm{\theta}^{(n_2)}_{k,+}) - f(\bm{\theta}^{(n_1)}_{k,-}),
\end{align*}
where  $\bm{\theta}^{(n_1)}_{k,-} = \bm{\theta}_k - \sum_{i=0}^{n_1-1} \frac{\gamma\nabla f(\bm{\theta}^{(i)}_{k,-})}{\norm{\nabla f(\bm{\theta}^{(i)}_{k,-})}}$, $\bm{\theta}^{(n_2)}_{k,+} = \bm{\theta}_k + \sum_{i=0}^{n_2-1} \gamma\frac{\nabla f(\bm{\theta}^{(i)}_{k,+})}{\norm{\nabla f(\bm{\theta}^{(i)}_{k,+})}}$ and $\bm{\theta}^{(0)}_{k,+} = \bm{\theta}^{(0)}_{k,-} =\bm{\theta}_k$.
\end{definition}


\begin{remark}
In contrast to the measures defined in~\eqref{eq:sharp_2} and~\eqref{eq:sharp_keskar}, Definition~\ref{def:sharp_approx} does not require a ball radius $\varepsilon$. However, our definition requires specifying the step-size $\gamma$ and the number of ascent and descent iterations.
\end{remark}
\begin{remark}
Running gradient descent/ascent with fixed step-size near a minimizer can return a very small sharpness value even if the minimizer is sharp. This is due to the small gradient norm around a minimizer. To resolve this issue, we normalize the gradient at every descent/ascent step. Moreover, normalizing by the norm of the gradient helps in understanding the radius of the ball containing the iterates $\{\btheta_{k,-}^{(j)}\}_{j=1}^{n_1}$ and $\{\btheta_{k,+}^{(j)}\}_{j=1}^{n_2}$.
\end{remark}

\begin{remark}
According to definition~\ref{def:sharp_approx}, larger gradient magnitudes yield larger sharpness values. Adam, a very popular method for training neural networks, tend to decrease the learning rate when the accumulated gradients are large; i.e., according to our definition of sharpness, the method returns reduced learning rate in sharp regions. This can be one explanation for the common conception that Adam converges faster than SGD, but generalizes worse \citep{keskar2017improving, zhou2020towards}. Our proposed method adapts a reverse behavior. More specifically, we aim at choosing a large learning rate in sharp regions. Our experiments indicate that this approach helps finding flat local minima. Interestingly, our experiments further show that SALR can improve the performance of Adam when adopting our learning rate schedule strategy.

\end{remark}

\begin{figure*}[!ht]
    \centering
     \centerline{\includegraphics[width=\columnwidth]{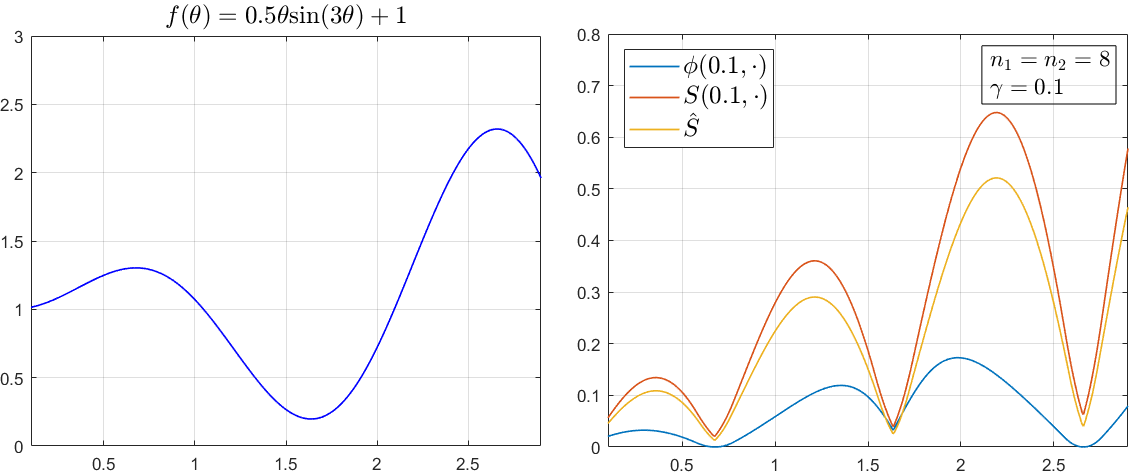}}
    \caption{Sharpness measure plots for $\phi$, $S$, and $\widehat{S}$ on function $f$.}
    \label{fig:sharpness_measure}
\end{figure*}

Figure~\ref{fig:sharpness_measure} shows the plots of the three different sharpness measures defined in this section when computed for a function $f(\theta) = 0.5\theta \, \mbox{sin}(3\theta) + 1$. Notice that the blue plot corresponding to the sharpness measure $\phi(\cdot)$ attains a zero value at local maximizers compared to a positive value for the other two sharpness plots. Moreover, notice that the sharpness value in these three plots attains a small value near the local minimizer. This can be explained by our choice of radius $\varepsilon = 0.1$ which limits the neighborhood being exploited. Increasing the radius for $\phi(\varepsilon, \cdot)$ and $S(\varepsilon, \cdot)$ (increasing $n_1$ and $n_2$ for $\widehat{S}$) will provide higher values around the minimizer. We next show that using our sharpness measure in Definition~\ref{def:sharp_approx}, gradient descent with SALR framework in Algorithm~\ref{algo:0}, denoted as GD-SALR, escapes sharp local minima. 

\section{Theoretical Results} \label{sec:theory}
In this section, we focus on analyzing the convergence of vanilla GD when adopting our sharpness-aware learning rate framework in Algorithm \ref{algo:0}. We show that GD-SALR escapes any given neighborhood of a sharp local minimum by choosing a sufficiently large step-size. Throughout this section, we make the following assumptions that are standard for the convergence theory of gradient descent methods.

\textbf{Assumptions}: The objective function $f$ is twice continuously differentiable and $L_0$-Lipschitz continuous. The gradient function $\nabla f(\cdot)$ is L-Lipschitz continuous with Lipschitz
constant $L$. Furthermore, the gradient norm is bounded, i.e. there exists a scalar constant $g_{max} > 0$ such that $\|\nabla f(\btheta)\| \leq g_{max}$ for all $\btheta$.

The next theorem shows that GD with a sufficiently large step-size escapes a given strongly convex neighborhood around a local minimum $\btheta^*$. The proof of the theorem is relegated to the Appendix.

\begin{theorem}\label{thm:lm-GD}
Suppose that $f$ is $\mu$-strongly convex function in a neighborhood $\ball_{\delta}(\btheta^*)$ around a local minimum $\btheta^*$, i.e. $\lambda_{min}\left(\nabla^2 f(\btheta) \right) \geq \mu$ for all $\btheta \in \ball_{\delta}(\btheta^*) \triangleq \{\btheta \, |\, \|\btheta - \btheta^*\|_2 \leq \delta\} $.  Running vanilla GD with $\btheta_0 \in \ball_{\delta}(\btheta^*)$  and learning rate $\eta_k \geq \dfrac{2+ \varepsilon}{\mu}$ for some fixed $\varepsilon > 0$, there exists $\widehat{k}$ with $\btheta_{\widehat{k}} \notin \ball_{\delta}(\btheta^*)$.
\end{theorem}


Our next result shows that GD-SALR escapes sharp local minima by dynamically choosing a sufficiently large step-size in a local strongly convex region. The proof is relegated to the Appendix.

\begin{theorem}\label{thm:GD_SALR}
Suppose that $f$ is a $\mu$-strongly convex function in a neighborhood $\ball_{\delta}(\btheta^*)$ around a local minimum $\btheta^*$ , i.e. $\lambda_{min}\left(\nabla^2 f(\btheta) \right) \geq \mu$ for all $\btheta \in \ball_{\delta}(\btheta^*) \triangleq \{\btheta \, |\, \|\btheta - \btheta^*\|_2 \leq \delta\} $. Under our Assumptions, run GD-SALR (Gradient descent with step size choice according to Algorithm~\ref{algo:0}) and Definition~\ref{def:sharp_approx}) with 
\begin{align*}
    &n_1 \geq \dfrac{a_1}{\left(\log\left( 1 + \dfrac{\mu\,g_{min}}{L\,g_{max} - \mu \, g_{min}}\right)\right)^2} - \dfrac{1}{a_1},\\
    &n_2 \geq\dfrac{a_2}{\left(\log\left( 1 + \dfrac{\mu\,g_{min}}{L\,g_{max}}\right)\right)^2} - \dfrac{1}{a_2}, \quad \mbox{and} \quad \gamma = \dfrac{g_{min}}{L},
\end{align*}
where 
\begin{align*}
    &a_1 = \dfrac{(2+\epsilon)L_0}{\eta_0(g_{max}L - g_{min} \mu)} + \dfrac{g_{min}\mu}{2(g_{max}L - g_{min} \mu)}, \quad \mbox{and} \\
    &  a_2 = \dfrac{(2+\epsilon)L_0}{\eta_0g_{max}L} + \dfrac{\mu^2 \, g_{min}}{2L^2g_{max}}, \quad  \epsilon, \eta_0 > 0, 
\end{align*}
and $g_{min}>0$ is a lower bound that satisfies 
\[\max\left\{\left\| \nabla f\left( \btheta_{k,-}^{(n_1-1)} \right)\right\|, \left\| \nabla f\left( \btheta_{k} \right) \right\|\right\} \geq g_{min}.\]
If $\delta > \max\{n_1, n_2\}\gamma$, then there exists $\widehat{k}$ with $\btheta_{\widehat{k}} \notin \ball_{\delta}(\btheta^*)$.
\end{theorem}
\begin{remark}
Our theorem states that when the function is strongly convex around a local minimizers (i.e. local minimizer is sharp), GD-SALR can escape the neighborhood by choosing a large enough step-size. Higher $\mu$ values (strong convexity parameter) reflects sharper minimizers. Our result shows that as $\mu$  increases, we require lower descent and ascent steps to escape the neighborhood. Notice that our result is local and does not require the convexity of the objective.

\end{remark}

\begin{remark}
In the context of machine learning, our theorem shows that our algorithm can potentially escape sharp regions even when all the data are used (full-batch). For instance, \citep{patel2017impact} shows that when using large batch sizes we require a higher learning rate to escape sharp minima. This provides an insight on the favorable empirical results presented in Section~\ref{sec:experiments} when running SGD-SALR. Moreover, our dynamic choice of high learning rates in sharp regions can potentially allow running SGD with larger batch sizes while still escaping sharp minimizers. This in turn provides an avenue for improved parallelism \citep{dean2012large, das2016distributed}. 

\end{remark} 

\section{Stochastic Approximation of Sharpness} \label{sec:stochastic}


\label{sec:back}
The concept of generalization is more relevant when solving problems arising in machine learning settings. Under the empirical risk minimization framework, the problem of training machine learning models can be mathematically formulated as the following optimization problem
\begin{equation}
    \min_{\bm{\theta}\in\mathbb{R}^n}f(\bm{\theta})\triangleq \frac{1}{m}\sum_{i=1}^m f_i(\bm{\theta}),
\end{equation}
where $f_i$ is a loss function parameterized with parameter $\btheta$ corresponding to data point $i \in \{1, 2, \ldots , m \}$. The most popular algorithm used to solve such optimization problems is stochastic gradient descent (SGD) which iteratively updates the parameters using the following update rule: 
\begin{align*}
    \bm{\theta}_{k+1}=\bm{\theta}_k-\eta_k\bigg(\frac{1}{|B_k|}\sum_{i\in B_k}\nabla f_i(\bm{\theta}_k)\bigg),
\end{align*}
where $B_k$ is the batch sampled at iteration $k$ and $\eta_k$ is the learning rate. To apply our framework in stochastic settings, we provide a stochastic procedure for computing the sharpness measure at a given iterate. Details are provided in Algorithm 2. By adopting Algorithm 2, our framework can be applied to numerous popular algorithms like SGD, Adam and Entropy-SGD.


\begin{algorithm}[!htbp]
    \caption{Calculating stochastic sharpness at iteration $k$}
    \SetAlgoLined
	\KwData{batch size $B_k$, base learning rate $\gamma$, current iterate $\bm{\theta}_k$, iteration number $n_1, n_2$ }
	Set $\bm{\theta}_{k,+}^{(0)}=\bm{\theta}_{k,-}^{(0)}=\bm{\theta}_k$\;
	\For{$i=0:n_1-1$}{
	    $\bm{\theta}_{k,-}^{(i+1)}=\bm{\theta}_{k,-}^{(i)}-\gamma\left(\dfrac{1}{|B_k|}\displaystyle{\sum_{j\in B_k}\dfrac{\nabla f_j\left(\bm{\theta}_{k,-}^{(i)})\right)}{\norm{\nabla f_j\left(\bm{\theta}_{k,-}^{(i)}\right)}}}\right)$\;
	}
	\For{$i=0:n_2-1$}{
	    $\bm{\theta}_{k,+}^{(i+1)}=\bm{\theta}_{k,+}^{(i)}+\gamma\left(\dfrac{1}{|B_k|}\displaystyle{\sum_{j\in B_k}\dfrac{\nabla f_j\left(\bm{\theta}_{k,+}^{(i)}\right)}{\norm{\nabla f_j \left(\bm{\theta}_{k,+}^{(i)}\right)}}}\right)$\;
	}
	Set $\hat S_{k} = \dfrac{1}{|B_k|}\displaystyle{\sum_{j\in B_k}f_j\left(\bm{\theta}_{k,+}^{(n_2)}\right)} - \dfrac{1}{|B_k|}\displaystyle{\sum_{j\in B_k}f_j\left(\bm{\theta}_{k,-}^{(n_1)}\right)}$\;
	    \label{algo:2}
\end{algorithm}

Here, we provide the detailed implementations of SGD-SALR and ADAM-SALR.

\begin{algorithm}[htb]
    \label{algo:sgd_salr}
    \small
	\SetAlgoLined
	\KwData{base learning rate $\eta_0$, number of iterations $K$, frequency $c$, initial weight $\bm{\theta}_0$}
	Set $\mathcal{S}=\emptyset$\;
    \For{$k=0:K$}{
        \If {$k\mod c=0$}{ Calculate $\hSk$ using Algorithm 2\; }
        Compute ${\cal S} = \mbox{Median}(\{\hSk\})$\;
        Set $\eta_k=\eta_0\dfrac{\hSk}{{\cal S}}$\;
        $\bm{\theta}_{k+1}=\bm{\theta}_k-\eta_k\dfrac{1}{|B_k|}\sum_{j\in B_k}\nabla f_j(\bm{\theta}_k)$\;
    }
    Set $\bm{\theta}=\bm{\theta}_K$\;
	Return $\bm{\theta}$\;
	\caption{The SGD-SALR Algorithm}
\end{algorithm}

\begin{algorithm}[htb]
    \label{algo:adam_salr}
	\SetAlgoLined
	\KwData{base learning rate $\eta_0$, exponential decay rates for the moment estimates $\beta_1,\beta_2\in[0,1)$, number of iterations $K$, frequency $c$, initial weight $\bm{\theta}_0$, perturbation $\epsilon$}
	\KwResult{weight vector $\bm{\theta}$.}
	Set $\mathcal{S}=\emptyset$\;
	Set $m_0=v_0=0$\;
    \For{$k=0:K$}{
        \If {$k\mod c=0$} {Calculate $\hSk$\;}
          Compute ${\cal S} =\mbox{Median}(\{\hSk\})$\\
        Set $\eta_k=\eta_0\dfrac{\hat S_{k}}{{\cal S}}$\;
        $g=\displaystyle{\dfrac{1}{|B_k|}\sum_{j\in B_k}\nabla f_j(\bm{\theta}_k)}$\\
        $m_{k+1}=\beta_1m_k+(1-\beta_1)g$\\
        $v_{k+1}=\beta_2v_k+(1-\beta_2)g^2$\\
        $\widehat m_{k+1}=m_k/(1-\beta_1^{k+1})$\\
        $\widehat v_{k+1}=v_k/(1-\beta_2^{k+1})$\\
        $\bm{\theta}_{k+1}=\bm{\theta}_k-\eta_k\widehat m_{k+1}/(\sqrt{\widehat v_{k+1}}+\epsilon)$\;
    }
    Set $\bm{\theta}=\bm{\theta}_K$\;
	Return $\bm{\theta}$.
	\caption{The ADAM-SALR Algorithm}
\end{algorithm}

\section{Empirical Results} 
\label{sec:experiments}

In this section, we present experimental results on image classification, text prediction and finetuning tasks. We show that our framework SALR can be adopted by many optimization methods and achieve notable improvements over a broad range of networks. We compare SALR with Entropy-SGD \citep{chaudhari2019entropy}, SWA \citep{izmailov2018averaging} and SAM \citep{foret2021sharpness}. Besides those benchmarks, we also use the conventional SGD and Adam \citep{kingma2014adam} as baseline references. All aforementioned methods are trained with batch normalization \citep{ioffe2015batch} and dropout of probability 0.5 after each layer \citep{gal2016dropout}. We replicate each experiment 10 times to obtain the mean and standard deviation of testing errors. Moreover, to account for the overhead computation of our proposed method, we ensure all models have the \textit{same total number of gradient calls}. We consider some typical networks such as WideResNet \citep{zagoruyko2016wide}, ResNet \citep{he2016deep}, DenseNet \citep{iandola2014densenet}, MobileNetV2 \citep{sandler2018mobilenetv2} and RegNetX \citep{radosavovic2020designing}.

\subsection{Image Classification}
\label{subsec:image}
\subsubsection{CIFAR-10/100}
We run our proposed SGD-SALR method detailed in Algorithm 3 for 40 epochs. We collect the sharpness measure every $c=2$ iterations and set $n_1=n_2=5$. The experimental settings for other benchmark models are as follows: (1) \textbf{SGD:} we run SGD for 200 epochs using decay learning rates. (2) \textbf{SWA:} the setting is the same as SGD. In the SWA stage, we switch to a cyclic learning rate schedule as suggested in \citep{izmailov2018averaging}. (3) \textbf{Entropy-SGD:} following the setting in \citep{chaudhari2019entropy}, we train Entropy-SGD for 40 epochs and set Langevin iterations $L_a=5$. (4) \textbf{Entropy-SGD-SALR:} the setting is same as Entropy-SGD, however, we update the learning rate of Entropy-SGD using Algorithm 1. (5) \textbf{SAM} \citep{foret2021sharpness}: we run SAM for 100 epochs (in SAM, each iteration has 2 gradient calls). The choice of hyperparameter follows the guideline in the SAM paper. The results are reported in Table \ref{table:cifar10}. Overall, all methods have the same number of gradient calls (i.e., wall-clock times).

\begin{table*}[!htbp]
\caption{Classification accuracy on CIFAR10}
\small
\centering
\begin{tabular}{ccccccc}
\hline
\textbf{Network} & \textbf{SGD} & \textbf{SWA} & \makecell{\textbf{Entropy-SGD}}  & \makecell{\textbf{Entropy-SGD-} \\ \textbf{SALR}} & \textbf{SAM} & \makecell{\textbf{SGD-SALR}}  \\ \hline
ResNet50 & 93.25 (0.03) & 93.31 (0.06) & 93.77 (0.08)  & 94.47 (0.12) & 94.50 (0.08) & \textbf{94.94} (0.09)    \\ \hline
All-CNN-BN & 91.93 (0.01) & 92.20 (0.01) & 91.13 (0.01) & 92.16 (0.01) & 92.18 (0.03) & \textbf{92.45} (0.05)    \\ \hline
ResNet101 & 95.11 (0.02) & 95.56 (0.01) & 95.51 (0.01) & 95.87 (0.01) & 95.92 (0.03) & \textbf{95.99} (0.00)    \\ \hline
RegNetX & 94.24 (0.02) & 94.23 (0.01) & 94.26 (0.01) & 95.00 (0.01) & 94.39 (0.05) & \textbf{95.01} (0.01)    \\ \hline
\hline
\textbf{Network} & \textbf{Adam} & \textbf{SWA} & \makecell{\textbf{Entropy-Adam}}  & \makecell{\textbf{Entropy-Adam-} \\ \textbf{SALR}} &  \textbf{SAM} &\makecell{\textbf{Adam-SALR}} \\ \hline
ResNet50 & 92.91 (0.07) & 92.43 (0.06) & 92.61 (0.11)  & 93.15 (0.09) & 93.27 (0.10) & \textbf{93.61} (0.09)    \\ \hline
All-CNN-BN & 91.95 (0.01) & 92.27 (0.01) & 91.10 (0.01) & 92.15 (0.01) & 92.13 (0.03) & \textbf{92.35} (0.01)    \\ \hline
ResNet101 & 95.00 (0.01) & 95.57 (0.01) & 95.56 (0.01) & 96.03 (0.01) & \textbf{96.17} (0.08) & 95.97 (0.00)    \\ \hline
RegNetX & 94.33 (0.01) & 94.12 (0.01) & 94.21 (0.01) & \textbf{95.06} (0.01) & 95.04 (0.03) & 95.02 (0.01)    \\ \hline
\end{tabular}
\label{table:cifar10}
\end{table*}

\begin{table*}[!htbp]
\caption{Sharpness of final solutions (CIFAR-10, SGD)}
\small
\centering
\begin{tabular}{ccccccc}
\hline
$\times 10^{-3}$ & \textbf{SGD} & \textbf{SWA} & \makecell{\textbf{Entropy-SGD}}  & \makecell{\textbf{Entropy-SGD} \\ \textbf{SALR}} & \textbf{SAM} & \makecell{\textbf{SGD-SALR}}  \\ \hline
ResNet50 & 8.19 (0.50) & 7.51 (0.31) & 3.55 (0.47)  & 3.67 (0.28) & 3.45 (0.51) &\textbf{3.22} (0.63)    \\ \hline
All-CNN-BN & 11.02 (1.00) & 10.65 (1.21) & \textbf{6.12} (0.88) & 6.35 (0.84) & 6.34 (0.55) &6.30 (0.91)    \\ \hline
ResNet101 & 7.00 (0.87) & 6.91 (0.22) & 5.98 (0.60) & 6.07 (0.82) & 6.09 (0.65) & \textbf{5.53} (0.70)    \\ \hline
RegNetX & 9.56 (1.03) & 9.66 (0.69) & 9.41 (0.50) & 8.77 (0.68) & 8.69 (0.56) & \textbf{8.50} (0.71)    \\ \hline
\end{tabular}
\label{table:sharp1}
\end{table*}

To illustrate the flexibility of our framework, we change the base optimizer SGD to Adam and re-run all the experiments under a similar setting as that of Table \ref{table:cifar10}. Results are reported in the same Table. Furthermore, in Table \ref{table:sharp1}, we report the sharpness measure of the final solution obtained by each optimization approach.

Under a similar setting, we test all models on CIFAR-100. The batch size is 256. Experimental results are reported in Table \ref{table:cifar100}. In this experiment, all methods also run with the same number of gradient calls. 

\begin{table*}[htb]
\caption{Classification accuracy on CIFAR-100}
\normalsize
\label{table:cifar100}
\centering
\begin{tabular}{cccccc}
\hline
\textbf{Network} & \textbf{SGD} & \textbf{SWA} & \textbf{SAM} & \makecell{\textbf{Entropy-SGD}}  & \makecell{\textbf{SGD-SALR}} \\ \hline
ResNet18 & 79.20 (0.13) & 79.63 (0.08) & 81.30 (0.10) & 81.17 (0.12) & \textbf{81.73} (0.10) \\ \hline
RegNetX & 79.42 (0.05) & 79.45 (0.08) & 81.55 (0.09) & 81.61 (0.10) & \textbf{82.00} (0.13) \\ \hline
MobileNetV2 & 81.22 (0.06) & 81.18 (0.09) & 82.07 (0.06) & \textbf{82.53} (0.05) & 82.47 (0.09)    \\ \hline
WideResNet-28-10 & 82.00 (0.10) & 82.33 (0.09) & \textbf{83.52} (0.19) & 82.37 (0.12) & 83.12 (0.09)  \\ \hline
PyramidNet & 80.02 (0.08) & 80.31 (0.09) & 85.31 (0.17) & 84.88 (0.13) & \textbf{85.69} (0.10) \\ \hline
\end{tabular}
\end{table*}

\subsubsection{ImageNet}
Following the procedures in \citep{he2016deep, szegedy2016rethinking}, we train SALR and all benchmark models on ImageNet using ResNet152 and DenseNet161. The batch size is chosen to be 4096. The best testing accuracies are reported in Table \ref{table:2}. Furthermore, we increase the number of training epochs and report results in the same Table.

\begin{table*}[!htbp]
\caption{Classification accuracy on ImageNet}
\normalsize
\centering
\begin{tabular}{cccccc}
\hline
\textbf{Network} & \makecell{\textbf{SGD}\\ \textbf{200 epochs}} & \makecell{\textbf{SWA}\\ \textbf{200 epochs}} & \makecell{\textbf{SAM}\\\textbf{100 epochs}} & \makecell{\textbf{Entropy-SGD} \\ \textbf{40 epochs}} & \makecell{\textbf{SGD-SALR} \\ \textbf{40 epochs}} \\ \hline
ResNet152 & 78.83 (0.08) & 79.21 (0.09) & 80.00 (0.11) & 79.44 (0.12) & \textbf{80.33 (0.10)} \\ \hline
DenseNet161 & 77.61 (0.07) & 78.24 (0.13) & 78.88 (0.07) & 78.90 (0.10) & \textbf{79.10 (0.07)} \\ \hline
\hline
\textbf{Network} & \makecell{\textbf{SGD}\\ \textbf{500 epochs}} & \makecell{\textbf{SWA}\\ \textbf{500 epochs}} & \makecell{\textbf{SAM}\\\textbf{250 epochs}} & \makecell{\textbf{Entropy-SGD} \\ \textbf{100 epochs}} & \makecell{\textbf{SGD-SALR} \\ \textbf{100 epochs}} \\ \hline
ResNet152 & 78.20 (0.06) & 78.21 (0.06) & 80.89 (0.11) & 79.94 (0.11) & \textbf{81.15 (0.09)} \\ \hline
DenseNet161 & 77.13 (0.12) & 77.21 (0.10) & 79.25 (0.10) & 78.99 (0.10) & \textbf{79.37 (0.11)} \\ \hline
\hline
\textbf{Network} & \makecell{\textbf{SGD}\\ \textbf{800 epochs}} & \makecell{\textbf{SWA}\\ \textbf{800 epochs}} & \makecell{\textbf{SAM}\\\textbf{400 epochs}} & \makecell{\textbf{Entropy-SGD} \\ \textbf{160 epochs}} & \makecell{\textbf{SGD-SALR} \\ \textbf{160 epochs}} \\ \hline
ResNet152 & 78.03 (0.01) & 78.08 (0.00) & 81.59 (0.01) & 81.05 (0.02) & \textbf{81.70} (0.03) \\ \hline
DenseNet161 & 77.00 (0.03) & 77.01 (0.02) & 79.33 (0.03) & \textbf{79.43} (0.01) & 79.40 (0.01) \\ \hline
\end{tabular}
\label{table:2}
\end{table*}

As seen in Table \ref{table:2}, SALR can improve performance of SGD even on the complex image classification task. Notably, as we increase the number of training epochs, the performance of SGD and SWA decrease due to model overfitting. However, SAM, Entropy-SGD and SGD-SALR do not suffer from this issue. This further demonstrates the advantage of sharpness-aware algorithms.

\subsection{Text Prediction}
We train an LSTM network on the Penn Tree Bank (PTB) dataset for word-level text prediction. This dataset contains about one million words. Following the guideline in [1] and [2], we train PTB-LSTM with 66 million weights. SGD and SWA are trained with 55 epochs. Entropy-SGD ($L=5$) and SALR ($c=2$) are trained with 11 epochs. SAM is trained with 28 epochs. Overall, all methods have the same number of gradient calls.

Our second task is to train an LSTM to perform character-level text-prediction using War and Peace (WP). We follow the procedures in [2] and [3]. We train SGD/SWA and Entropy-SGD/SALR with 50 and 10 epochs, respectively. We report the perplexity on the test set in Table \ref{table:per}. 

    \begin{table*}[htb]
        \caption{Perplexity on PTB/WP}
        \normalsize
    \centering
    \begin{tabular}{cccccc}
    \hline
    \textbf{PTB} & \textbf{SGD} & \textbf{SWA} & \textbf{Entropy-SGD} & \textbf{SAM} & \textbf{SGD-SALR} \\ \hline
    PTB-LSTM     & 78.4 (0.22)  & 78.1 (0.25)  & 72.15 (0.16)     & 72.10 (0.18)  & \textbf{71.42 (0.14)}  \\ \hline
    WP-LSTM     & 1.223 (0.01)  & 1.220 (0.05)  & 1.095 (0.01)    & 1.091 (0.02)    & \textbf{1.089 (0.02)}  \\ \hline
    \end{tabular}
    \label{table:per}
    \end{table*}

    \begin{table*}[!htbp]
    \caption{Top-1 testing accuracy for finetuning tasks.}
    \normalsize
    \centering
    \begin{tabular}{ccccc}
    \hline
    \textbf{EffNet-b7} & \textbf{SGD} & \makecell{\textbf{Entropy-SGD}} & \textbf{SAM} & \makecell{\textbf{SGD-SALR}}  \\ \hline
    Flowers & 98.83 (0.03) & 99.01 (0.02) & \textbf{99.37} (0.02)  & 99.35 (0.01) \\ \hline
    CIFAR-100 & 92.31 (0.06) & 92.49 (0.05) & 92.56 (0.06) & \textbf{92.60} (0.05)  \\ \hline
    Birdsnap & 85.71 (0.17) & 86.10 (0.15) & 86.36 (0.18) & \textbf{86.38} (0.12)  \\ \hline
    ImageNet & 84.69 (0.02) & 84.88 (0.03) & 84.86 (0.02) & \textbf{84.93} (0.02) \\ \hline
    \\ \hline
    \textbf{EffNet-L2} & \textbf{SGD} & \makecell{\textbf{Entropy-SGD}} & \textbf{SAM} & \makecell{\textbf{SGD-SALR}}  \\ \hline
    FGVC-Aircraft & 94.21 (0.07) & 94.74 (0.05) & 95.18 (0.06) & \textbf{95.41} (0.05) \\ \hline
    Food101 & 96.00 (0.03) & \textbf{96.30} (0.02) & 96.20 (0.01) & 96.24 (0.01) \\ \hline
    Birdsnap & 89.70 (0.17) & 89.64 (0.13) & \textbf{90.07} (0.14) & 90.05 (0.11) \\ \hline
    ImageNet & 88.20 (0.05) & 88.33 (0.03)  & 88.62 (0.03) & \textbf{88.75} (0.05) \\ \hline
    \end{tabular}
    \label{table:fine}
    \end{table*}

\begin{figure*}[!htbp]
    \caption{(Left) All-CNN-BN: Change of Testing Errors over Epochs. (Right) Change of sharpness and learning rate over iterations.}
    \centering
    \centerline{\includegraphics[width=\columnwidth]{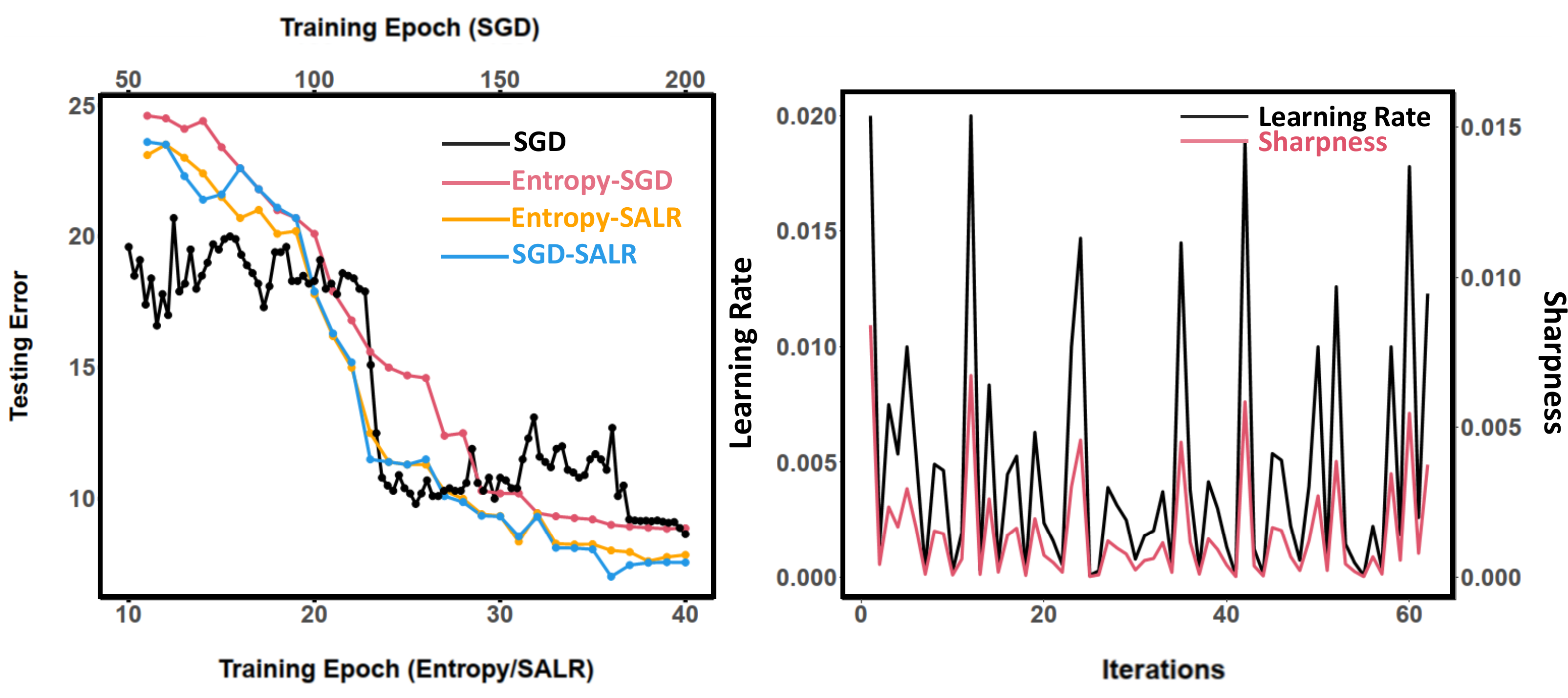}}
    \label{fig:acc}
\end{figure*}

\subsection{Fine-tuning}
We test our algorithm and other benchmarks on several finetuning tasks. We use EfficientNet-b7 and EfficientNet-L2 pretrained on ImageNet. Weight parameters are initialized to the values provided by the publicly available checkpoints \citep{tan2019efficientnet} while a new output layer with random weights is added to accommodate labels from a new dataset. Following the guideline in \citep{foret2021sharpness}, we use $\eta_0=0.016$, batch size of 1024 with weight decay $1e^{-5}$. SAM is trained with 5k steps, and Entropy-SGD and SGD-SALR are trained with 2k steps. For ImageNet, we train SAM with 10 epochs and train Entropy-SGD/SGD-SALR with 4 epochs. The initial learning rate is set to be 0.1. Results are reported in Table \ref{table:fine}.

As shown in Table \ref{table:fine}, Entropy-SGD, SAM and SGD-SALR can consistently improve the performance of SGD on several high-performance finetuning tasks. This further demonstrate the advantage of sharpness-aware algorithm.

\subsection{Additional Analysis}

\begin{table*}[!htbp]
\caption{Sensitivity analysis on $\gamma$}
\normalsize
\centering
\begin{tabular}{ccccccc}
\hline
$\gamma$ & $5\times 10^{-4}$ & $1\times 10^{-3}$ & $2\times 10^{-3}$ & $5\times 10^{-3}$ & 0.01 & 0.02 \\ \hline
Accuracy & 85.63 (0.08) & 85.69 (0.10) & 85.45 (0.07) & 85.72 (0.03) & 85.68 (0.04) & 85.27 (0.05) \\ \hline
Final Sharpness ($10^{-3}$) & 4.87 (0.29) & 4.22 (0.31) & 5.15 (0.28) & 3.89 (0.18) & 4.34 (0.17) & 5.66 (0.20)  \\ \hline
\end{tabular}
\label{table:sharp}
\end{table*}

Based on all aforementioned tables, we can obtain some important insights. First, methods adopting SALR show superior performance over their benchmarks. This increase in classification accuracy (or decrease in perplexity) is consistent across various datasets and a range of network structures. We also observed that SGD-SALR tends to outperform other SALR based methods in most settings while achieving comparable results in others. Second, and more interestingly, this superior performance is achieved with 5 times less epochs compared to SGD, Adam and SWA. The caveat however is that SALR, Entropy and SAM respectively require $(n_1+n_2)/c=5$, $L_a=5$ and $2$ times more gradient calls at each iteration, hence making the total computational needs the same as Adam, SGD and SWA. Third and as shown in Table \ref{table:sharp}, it is clear that SALR drives solutions to significantly flatter regions. This highlights the effectiveness of dynamically adjusting learning rates based on the relative sharpness of the current iterate. To further demonstrate the advantage of SALR framework, we plot the testing error curves for SGD, Entropy-SGD, Entropy-SALR and SALR in Figure \ref{fig:acc} (Left). Interestingly, we can observe a smoother convergence when adopting SALR framework. To verify the behavior of our framework, we plot in Figure \ref{fig:acc} the dynamics of both the learning rate and sharpness when adopting SALR framework. The desired behavior can be clearly seen in this figure which shows a proportional relationship between learning rates and sharpness.

\subsection{Sensitivity Analysis}
To further study the effect of $\gamma$, we run sensitivity experiments on CIFAR-100 using PyramidNet (5 reps for each $\gamma$). The parameter $\gamma$ plays a role similar to $\epsilon$ in Definition \ref{def:sharp_approx}. More specifically, reducing this parameter limits the set of reachable points. This implies decreasing the ball radius. In practice, the most reasonable approach is to first specify the ball radius $\epsilon$, choose $\gamma$ that is of the same order of $\epsilon$ and run gradient methods. As shown in the Table \ref{table:sharp}, the impact of $\gamma$ is rather marginal on the final accuracy if it specified based on our recommendation. 



\section{Discussion \& Open Problems}
\label{sec:conclusion}
In this paper we introduce SALR: a sharpness-aware learning rate scheduler that aims to recover flat minima. To demonstrate the effective of SALR, we apply our framework over a wide range network structures for image classification, text prediction and finetuning tasks. Our empirical results indicate improved generalization performance when compared to SGD and many other benchmark optimization methods. For fair comparison, our results are achieved when the same number of gradient calls are used for different methods. 

Designing learning rate updating mechanisms that utilize the structure of the underlying landscape can potentially improve training deep models. Our research is one step forward in that direction. Further potential applications of such updating mechanism can be found in Bayesian deep learning~\citep{smith2017cyclical, zhang2019cyclical} and multi-modal MCMC inference. Through SALR, the cyclical schedule can be set through exploiting sharpness information. In lieu of this, SALR may also find use for multi-modal MCMC inference or defining stoppage criteria. A direct extension of our work can be seen in studying quasi-Newton approximations of the sharpness measure. Such methods can potentially provide more accurate approximations. 

\section*{Appendix}

\section{Proof of Theorem 5}\label{Appendix:Proof Theorem 4}
\begin{proof}
According to the update rule of vanilla gradient descent, it follows that by mean-value theorem there exists $\widehat{\btheta} \in [\btheta_k, \btheta^*]$ such that
\[\begin{array}{ll}
\btheta_{k+1} - \btheta^* &=  \btheta_{k} - \btheta^* - \eta_k \nabla f(\btheta_k) \\
                        &= \btheta_{k} - \btheta^* - \eta_k \left[\nabla f(\btheta^*) +  \nabla^2 f(\widehat{\btheta})(\btheta_{k} - \btheta^*) \right]\\
                        &= \left[\bI - \eta_k \,\nabla^2 f (\widehat{\btheta}) \right](\btheta_{k} - \btheta^*),
\end{array}\]
where the last equality holds since $\btheta^*$ is a local minimum. By taking the norm, we get
\begin{align*}
    \|\btheta_{k+1} - \btheta^* \|   &=  \left\|\left[\bI - \eta_k \,\nabla^2 f (\widehat{\btheta}) \right](\btheta_{k} - \btheta^*) \right\|\\
    &\geq  \left| 1 - \eta_k\,\mu  \right|\,  \|\btheta_{k} - \btheta^* \|,
\end{align*}
where the last inequality holds by our local strong convexity assumption, the fact that $\widehat{\btheta} \in \ball_{\delta}\{\btheta^*\}$ and our choice of $\eta_k$. The former choice also imply that
\[\|\btheta_{k+1}  - \btheta^*\| \geq(\eta_k \mu - 1) \|\btheta_k  - \btheta^* \| \geq (1 + \epsilon)\|\btheta_k  - \btheta^* \|,\]
which yields
\[\|\btheta_k  - \btheta^*\| \geq (1+\varepsilon)^k \|\btheta_0  - \btheta^*\|. \]
Let $\|\btheta_0 - \btheta^*\| = D$ and $\widehat{k} = \dfrac{1}{\log(1+\varepsilon)}\log\left( \dfrac{\delta}{D}\right)$, then 
\[\|\btheta_{\widehat{k}}  - \btheta^*\| \geq \delta,  \]
which completes our proof.

\end{proof}

\section{Proof of theorem 6}\label{Appendix:Proof Theorem 5}
\begin{proof}
According to Theorem 5, running vanilla gradient descent with $\eta_k \geq \dfrac{2+\varepsilon}{\mu}$ for some fixed $\varepsilon>0$ escapes the neighborhood $\ball_{\delta}(\btheta^*)$. Hence, to complete our proof, it suffices to show that GD-SALR will dynamically choose a sufficiently large step size.

We first start by computing a lower bound for our local sharpness approximation in local strongly convex regions. By definition, 
\[\hSk = f\left(\btheta_{k,+}^{(n_2)}\right)  - f\left(\btheta_{k,-}^{(n_1)}\right) = f\left(\btheta_{k,+}^{(n_2)}\right)  - f(\btheta_k) + f(\btheta_k) - f\left(\btheta_{k,-}^{(n_1)}\right).\]

We start by computing a lower bound for $f(\btheta_k) - f\left(\btheta_{k,-}^{(n_1)}\right)$. By descent lemma~\citep{bertsekas1997nonlinear},
\[\begin{array}{ll}
f\left(\btheta_{k,-}^{(i+1)}\right) & \leq  f\left(\btheta_{k,-}^{(i)}\right) + \left\langle \nabla f\left( \btheta_{k,-}^{(i)} \right) , \btheta_{k,-}^{(i+1)} - \btheta_{k,-}^{(i)} \right\rangle + \dfrac{L}{2}\left\| \btheta_{k,-}^{(i+1)} -\btheta_{k,-}^{(i)}  \right\|^2\\
& = f\left(\btheta_{k,-}^{(i)}\right) - \gamma  \left\| \nabla f\left( \btheta_{k,-}^{(i)} \right) \right\| +  \dfrac{L\gamma^2}{2}.
\end{array}\]

By summing over the $n_1$ iterations, we get
\begin{equation}\label{eq:LB_-_1}
f(\btheta_k) - f\left(\btheta_{k,-}^{(n_1)}\right) \geq \gamma \sum_{i=0}^{n_1-1} \left\|\nabla f\left( \btheta_{k,-}^{(i)} \right)  \right\| - \dfrac{n_1 L \gamma^2}{2}.
\end{equation}

By mean value theorem, there exists $\bz_{k,-}^{(i)} \in \left[\btheta_{k,-}^{(i)}, \btheta_{k,-}^{(i+1)} \right]$ with
\[\begin{array}{ll}
\nabla f\left( \btheta_{k,-}^{(i+1)} \right) &= \nabla f\left(\btheta_{k,-}^{(i)}\right) + \nabla^2 f\left(\bz_{k,-}^{(i)} \right)\left( \btheta_{k,-}^{(i+1)} - \btheta_{k,-}^{(i)} \right)\\
\\
&=\nabla f\left(\btheta_{k,-}^{(i)}\right) - \gamma \nabla^2 f\left(\bz_{k,-}^{(i)} \right) \dfrac{\nabla f\left( \btheta_{k,-}^{(i)}\right)}{\left\|\nabla f(\btheta_{k,-}^{(i)}  \right\|}
\end{array},\]
which yields
\[\left\| \nabla f\left(\btheta_{k,-}^{(i+1)} \right)  \right\| \leq \left( 1 - \gamma \mu/g_{max} \right) \left\| \nabla f\left( \btheta_{k,-}^{(i)} \right) \right\| = \left( \dfrac{L \,g_{max} -  \mu\,g_{min}}{L\,g_{max}} \right) \left\| \nabla f\left( \btheta_{k,-}^{(i)} \right) \right\| .\]
Here the inequality holds by local strong convexity and the fact that $\bz_{k,-}^i \in \ball_{\delta}(\btheta^*)$ due to our choice of $\delta$, the upper bound on the norm of the gradient, and our choice of $\gamma$. Substituting back into~\eqref{eq:LB_-_1}, we get
\begin{equation}\label{eq:LB_-_2}
\begin{array}{ll}
f(\btheta_k) - f\left(\btheta_{k,-}^{(n_1)}\right) 
&\geq \gamma \displaystyle{\sum_{i=0}^{n_1-1}\left(\dfrac{L \,g_{max} -  \mu\,g_{min}}{L\,g_{max}}\right)^{-i} \left\|\nabla f\left( \btheta_{k,-}^{(n_1-1)} \right)  \right\| - \dfrac{n_1 L \gamma^2}{2}}\\\\
&= \dfrac{g_{min}}{L} \left(\dfrac{L \, g_{max}}{\mu g_{min}} -1 \right) \left(\left(\dfrac{L\,g_{max}}{L \,g_{max} -  \mu\,g_{min}}\right)^{n_1} - 1 \right) \left\|\nabla f\left( \btheta_{k,-}^{(n_1-1)} \right)  \right\| - \dfrac{n_1 L \gamma^2}{2}\\\\
&= \left(\dfrac{g_{max}}{\mu} -\dfrac{g_{min}}{L} \right) \left(\left(\dfrac{L\,g_{max}}{L \,g_{max} -  \mu\,g_{min}}\right)^{n_1} - 1 \right) \left\|\nabla f\left( \btheta_{k,-}^{(n_1-1)} \right)  \right\| - \dfrac{n_1 L \gamma^2}{2}\\
\end{array}
\end{equation}

By our choice of $n_1$, we have
\begin{equation}\label{eq:LB_-_3}
\left(1 + \dfrac{\mu g_{min} }{L \, g_{max} - \mu\,g_{min}}\right)^{n_1} -1 \geq n_1 \,a_1.
\end{equation}
The inequality holds since
\[\log \left(1+\dfrac{\mu \, g_{min} }{L \, g_{max} - \mu \, g_{min}}\right)^{n_1} = n_1 \,\log \left(1+\dfrac{\mu \, g_{min} }{L \, g_{max} - \mu \, g_{min}}\right) \geq  \dfrac{n_1 \, a_1}{\sqrt{n_1\,a_1+1}} \geq \log(1 + n_1a_1),\]
where the first inequality holds by our choice of $n_1$ and the second inequality is an upper bound of $\log(1+x)$. By substituting~\eqref{eq:LB_-_3} in~\eqref{eq:LB_-_2} and using our assumption that
\[\min\left\{\left\| \nabla f\left( \btheta_{k,-}^{(n_1-1)} \right)\right\|, \left\| \nabla f\left( \btheta_{k,+}^{0} \right) \right\|\right\} \geq g_{min},\]
we get
\begin{equation}\label{eq:LB_-_Final}
\begin{array}{ll}
f(\btheta_k) - f\left(\btheta_{k,-}^{(n_1)}\right) & \geq 
\left(\dfrac{g_{max}}{\mu} -\dfrac{g_{min}}{L} \right) n_1\,a_1 \, g_{min} - \dfrac{n_1  \,g_{min}^2}{2 L}\\\\
&\geq  n_1\, \left(\dfrac{2 + \epsilon}{ \eta_0 \mu}\right) \, \left(\dfrac{L_0 \, g_{min}}{L}\right).
\end{array}
\end{equation}

We now compute a lower bound for $f\left(\btheta_{k,+}^{(n_2)}\right)  - f(\btheta_k)$. By local strong convexity of $f$, we have
\[\begin{array}{ll}
f\left(\btheta_{k,+}^{(i+1)}\right) & \geq  f\left(\btheta_{k,+}^{(i)}\right) + \left\langle \nabla f\left( \btheta_{k,+}^{(i)} \right) , \btheta_{k,+}^{(i+1)} - \btheta_{k,+}^{(i)} \right\rangle + \dfrac{\mu}{2}\left\| \btheta_{k,+}^{(i+1)} -\btheta_{k,+}^{(i)}  \right\|^2\\\\
& = f\left(\btheta_{k,+}^{(i)}\right) + \gamma  \left\| \nabla f\left( \btheta_{k,+}^{(i)} \right) \right\| +  \dfrac{\mu\gamma^2}{2}.
\end{array}\]

By summing over the $n_2$ iterations, we get
\begin{equation}\label{eq:LB_+}
f\left(\btheta_{k,+}^{(n_2)}\right) - f(\btheta_k) \geq \gamma \sum_{i=0}^{n_2-1} \left\|\nabla f\left( \btheta_{k,+}^i \right)  \right\| + \dfrac{n_2 \mu \gamma^2}{2}.
\end{equation}

By mean value theorem, there exists $\bz_{k,+}^i \in \left[\btheta_{k,+}^i, \btheta_{k,+}^{(i+1)} \right]$ with
\[\begin{array}{ll}
\nabla f\left( \btheta_{k,+}^{(i+1)} \right) &= \nabla f\left(\btheta_{k,+}^{(i)} \right) + \nabla^2 f\left(\bz_{k,+}^i \right)\left( \btheta_{k,+}^{(i+1)} - \btheta_{k,+}^{(i)} \right)\\
\\
&=\nabla f\left(\btheta_{k,+}^{(i)}\right) + \gamma \nabla^2 f\left(\bz_{k,+}^i \right) \dfrac{\nabla f\left( \btheta_{k,+}^{(i)}\right)}{\left\|\nabla f\left( \btheta_{k,+}^{(i)}\right) \right\|},
\end{array}\]
which yields
\[\left\| \nabla f\left(\btheta_{k,+}^{(i+1)} \right)  \right\| \geq \left( 1 + \gamma \mu / g_{max} \right) \left\| \nabla f\left( \btheta_{k,+}^{(i)} \right) \right\|.\]
Here the inequality holds by local strong convexity and the fact that $\bz_{k,+}^i \in \ball_{\delta}(\btheta^*)$, the upper bound on the norm of the gradient, and our choice of $\gamma$. Substituting back into~\eqref{eq:LB_+}, we get
\begin{equation}\label{eq:LB_+_2}
\begin{array}{ll}
 f\left(\btheta_{k,+}^{(n_2)}\right) - f(\btheta_k) &\geq \displaystyle{\gamma \sum_{i=0}^{n_2-1}\left(1 + \gamma \mu/g_{max} \right)^i \left\|\nabla f\left( \btheta_{k,+}^{0} \right)  \right\| + \dfrac{n_2 \mu \gamma^2}{2}}\\\\
&=  \dfrac{g_{max}}{\mu}\left(\left(1 + \dfrac{\mu g_{min} }{L \, g_{max}}\right)^{n_2} -1 \right) \left\|\nabla f\left( \btheta_{k,+}^{0} \right)  \right\| + \dfrac{n_2 \mu \gamma^2}{2}\\
\end{array}
\end{equation}

By our choice of $n_2$, we have
\begin{equation}\label{eq:LB_+_3}
\left(1 + \dfrac{\mu g_{min} }{L \, g_{max}}\right)^{n_2} -1 \geq n_2 \,a_2.
\end{equation}
The inequality holds since
\[ \log \left(1+\dfrac{\mu g_{min} }{L \, g_{max}}\right)^{n_2} = n_2\log \left(1+\dfrac{\mu g_{min} }{L \, g_{max}}\right) \geq  \dfrac{n_2 a_2}{\sqrt{n_2a_2+1}} \geq \log(1 + n_2 a_2),\]
where the first inequality holds by our choice of $n_2$ and the second inequality is an upper bound of $\log(1+x)$. By substituting~\eqref{eq:LB_+_3} in~\eqref{eq:LB_+_2} and using our assumption that
\[\min\left\{\left\| \nabla f\left( \btheta_{k,-}^{(n_1-1)} \right)\right\|, \left\| \nabla f\left( \btheta_{k,+}^{0} \right) \right\|\right\} \geq g_{min},\]
we get
\begin{equation}\label{eq:LB_+_Final}
\begin{array}{ll}
f\left(\btheta_{k,+}^{(n_2)}\right) - f(\btheta_k) & \geq \dfrac{g_{max} n_2 a_2 \, g_{min}}{\mu} + \dfrac{n_2 \mu g_{min}^2}{2 L^2}\\\\
& \geq  n_2\, \left(\dfrac{2 + \epsilon}{\mu \,\eta_0 }\right) \, \left(\dfrac{L_0 \, g_{min}}{L}\right).
\end{array}
\end{equation}

By adding~\eqref{eq:LB_-_Final} and~\eqref{eq:LB_+_Final}, we obtain
\begin{equation}\label{eq:LB_S}
\begin{array}{ll}
\hSk &\geq  (n_1 + n_2) \left(\dfrac{2 + \epsilon}{\mu \, \eta_0}\right) \, \left(\dfrac{L_0 \, g_{min}}{L}\right).
\end{array}
\end{equation}

We now provide an upper bound for Median$\left(\hSk\right)$. Using the Lipschitz property of function, we have
\begin{equation}\label{eq:UB_-}
    f\left(\btheta_{k} \right)  - f\left(\btheta_{k,-}^{(n_1)}\right) = \sum_{i=0}^{n_1-1} f\left(\btheta_{k,-}^{(i)} \right)  - f\left(\btheta_{k,-}^{(i+1)} \right) \leq  L_0\sum_{i=0}^{n_1-1}  \left\|\btheta_{k,-}^{(i)} - \btheta_{k,-}^{(i+1)}  \right\|  = n_1 L_0 \gamma.
\end{equation}
\begin{equation}\label{eq:UB_+}
      f\left(\btheta_{k,+}^{(n_2)}\right) - f\left(\btheta_{k} \right) = \sum_{i=0}^{n_2-1} f\left(\btheta_{k,+}^{(i+1)} \right)  - f\left(\btheta_{k,+}^{(i)} \right) \leq  L_0\sum_{i=0}^{n_2-1} \gamma \left\|\btheta_{k,+}^{(i+1)} - \btheta_{k,+}^{(i)}  \right\|  = n_2\, L_0 \gamma.   
\end{equation}
Combining~\eqref{eq:UB_-} and~\eqref{eq:UB_+}, we get
\begin{equation}\label{eq:UB_S}
    {\cal S} = \mbox{Median}\left(\hSk\right)  \leq (n_1+n_2)L_0\gamma = (n_1 + n_2) \dfrac{g_{min}L_0}{L}.
\end{equation}
According to the definition of our learning rate, combining~\eqref{eq:LB_S} and~\eqref{eq:UB_S} results in the following inequality
\begin{equation}\label{eq:LR_U}
      \begin{array}{ll}
      \eta_k & = \eta_0 \dfrac{{\cal S}_k}{{\cal S}} \geq  \dfrac{2+ \epsilon}{\mu}.
      \end{array}
\end{equation}
The proof is concluded using Theorem 5.
\end{proof}

\bibliography{mybib}

\begin{thebibliography}{}

\bibitem[Allen-Zhu et~al., 2019]{allen2019convergence}
Allen-Zhu, Z., Li, Y., and Song, Z. (2019).
\newblock A convergence theory for deep learning via over-parameterization.
\newblock In {\em International Conference on Machine Learning}, pages
  242--252. PMLR.

\bibitem[Baldassi et~al., 2016]{baldassi2016unreasonable}
Baldassi, C., Borgs, C., Chayes, J.~T., Ingrosso, A., Lucibello, C., Saglietti,
  L., and Zecchina, R. (2016).
\newblock Unreasonable effectiveness of learning neural networks: From
  accessible states and robust ensembles to basic algorithmic schemes.
\newblock {\em Proceedings of the National Academy of Sciences},
  113(48):E7655--E7662.

\bibitem[Baldassi et~al., 2015]{baldassi2015subdominant}
Baldassi, C., Ingrosso, A., Lucibello, C., Saglietti, L., and Zecchina, R.
  (2015).
\newblock Subdominant dense clusters allow for simple learning and high
  computational performance in neural networks with discrete synapses.
\newblock {\em Physical review letters}, 115(12):128101.

\bibitem[Bertsekas, 1997]{bertsekas1997nonlinear}
Bertsekas, D.~P. (1997).
\newblock Nonlinear programming.
\newblock {\em Journal of the Operational Research Society}, 48(3):334--334.

\bibitem[Bottou and Bousquet, 2008]{bottou2008tradeoffs}
Bottou, L. and Bousquet, O. (2008).
\newblock The tradeoffs of large scale learning.
\newblock In {\em Advances in neural information processing systems}, pages
  161--168.

\bibitem[Bottou et~al., 2018]{bottou2018optimization}
Bottou, L., Curtis, F.~E., and Nocedal, J. (2018).
\newblock Optimization methods for large-scale machine learning.
\newblock {\em Siam Review}, 60(2):223--311.

\bibitem[Bottou and Le~Cun, 2005]{bottou2005line}
Bottou, L. and Le~Cun, Y. (2005).
\newblock On-line learning for very large data sets.
\newblock {\em Applied stochastic models in business and industry},
  21(2):137--151.

\bibitem[Bousquet and Elisseeff, 2002]{bousquet2002stability}
Bousquet, O. and Elisseeff, A. (2002).
\newblock Stability and generalization.
\newblock {\em Journal of machine learning research}, 2(Mar):499--526.

\bibitem[Chaudhari et~al., 2019]{chaudhari2019entropy}
Chaudhari, P., Choromanska, A., Soatto, S., LeCun, Y., Baldassi, C., Borgs, C.,
  Chayes, J., Sagun, L., and Zecchina, R. (2019).
\newblock Entropy-sgd: Biasing gradient descent into wide valleys.
\newblock {\em Journal of Statistical Mechanics: Theory and Experiment},
  2019(12):124018.

\bibitem[Choromanska et~al., 2015]{choromanska2015loss}
Choromanska, A., Henaff, M., Mathieu, M., Arous, G.~B., and LeCun, Y. (2015).
\newblock The loss surfaces of multilayer networks.
\newblock In {\em Artificial intelligence and statistics}, pages 192--204.

\bibitem[Das et~al., 2016]{das2016distributed}
Das, D., Avancha, S., Mudigere, D., Vaidynathan, K., Sridharan, S., Kalamkar,
  D., Kaul, B., and Dubey, P. (2016).
\newblock Distributed deep learning using synchronous stochastic gradient
  descent.
\newblock {\em arXiv preprint arXiv:1602.06709}.

\bibitem[Dean et~al., 2012]{dean2012large}
Dean, J., Corrado, G., Monga, R., Chen, K., Devin, M., Mao, M., Ranzato, M.,
  Senior, A., Tucker, P., Yang, K., et~al. (2012).
\newblock Large scale distributed deep networks.
\newblock In {\em Advances in neural information processing systems}, pages
  1223--1231.

\bibitem[Dinh et~al., 2017]{dinh2017sharp}
Dinh, L., Pascanu, R., Bengio, S., and Bengio, Y. (2017).
\newblock Sharp minima can generalize for deep nets.
\newblock {\em International Conference on Machine Learning}.

\bibitem[Duchi et~al., 2011]{duchi2011adaptive}
Duchi, J., Hazan, E., and Singer, Y. (2011).
\newblock Adaptive subgradient methods for online learning and stochastic
  optimization.
\newblock {\em Journal of machine learning research}, 12(7).

\bibitem[Dziugaite and Roy, 2017]{dziugaite2017computing}
Dziugaite, G.~K. and Roy, D.~M. (2017).
\newblock Computing nonvacuous generalization bounds for deep (stochastic)
  neural networks with many more parameters than training data.
\newblock {\em arXiv preprint arXiv:1703.11008}.

\bibitem[Foret et~al., 2021]{foret2021sharpness}
Foret, P., Kleiner, A., Mobahi, H., and Neyshabur, B. (2021).
\newblock Sharpness-aware minimization for efficiently improving
  generalization.
\newblock {\em International Conference on Learning Representations}.

\bibitem[Gal and Ghahramani, 2016]{gal2016dropout}
Gal, Y. and Ghahramani, Z. (2016).
\newblock Dropout as a bayesian approximation: Representing model uncertainty
  in deep learning.
\newblock In {\em international conference on machine learning}, pages
  1050--1059.

\bibitem[Garipov et~al., 2018]{garipov2018loss}
Garipov, T., Izmailov, P., Podoprikhin, D., Vetrov, D.~P., and Wilson, A.~G.
  (2018).
\newblock Loss surfaces, mode connectivity, and fast ensembling of dnns.
\newblock In {\em Advances in Neural Information Processing Systems}, pages
  8789--8798.

\bibitem[Gonen and Shalev-Shwartz, 2017]{gonen2017fast}
Gonen, A. and Shalev-Shwartz, S. (2017).
\newblock Fast rates for empirical risk minimization of strict saddle problems.
\newblock {\em JMLR: Workshop and Conference Proceedings}.

\bibitem[Goyal et~al., 2017]{goyal2017accurate}
Goyal, P., Doll{\'a}r, P., Girshick, R., Noordhuis, P., Wesolowski, L., Kyrola,
  A., Tulloch, A., Jia, Y., and He, K. (2017).
\newblock Accurate, large minibatch sgd: Training imagenet in 1 hour.
\newblock {\em arXiv preprint arXiv:1706.02677}.

\bibitem[Hardt et~al., 2016]{hardt2016train}
Hardt, M., Recht, B., and Singer, Y. (2016).
\newblock Train faster, generalize better: Stability of stochastic gradient
  descent.
\newblock In {\em International Conference on Machine Learning}, pages
  1225--1234.

\bibitem[He et~al., 2016]{he2016deep}
He, K., Zhang, X., Ren, S., and Sun, J. (2016).
\newblock Deep residual learning for image recognition.
\newblock In {\em Proceedings of the IEEE conference on computer vision and
  pattern recognition}, pages 770--778.

\bibitem[Hochreiter and Schmidhuber, 1997]{hochreiter1997flat}
Hochreiter, S. and Schmidhuber, J. (1997).
\newblock Flat minima.
\newblock {\em Neural Computation}, 9(1):1--42.

\bibitem[Iandola et~al., 2014]{iandola2014densenet}
Iandola, F., Moskewicz, M., Karayev, S., Girshick, R., Darrell, T., and
  Keutzer, K. (2014).
\newblock Densenet: Implementing efficient convnet descriptor pyramids.
\newblock {\em arXiv preprint arXiv:1404.1869}.

\bibitem[Ioffe and Szegedy, 2015]{ioffe2015batch}
Ioffe, S. and Szegedy, C. (2015).
\newblock Batch normalization: Accelerating deep network training by reducing
  internal covariate shift.
\newblock {\em International Conference on International Conference on Machine
  Learning}.

\bibitem[Izmailov et~al., 2018]{izmailov2018averaging}
Izmailov, P., Podoprikhin, D., Garipov, T., Vetrov, D., and Wilson, A.~G.
  (2018).
\newblock Averaging weights leads to wider optima and better generalization.
\newblock {\em Uncertainty in Artificial Intelligence}.

\bibitem[Kawaguchi, 2016]{kawaguchi2016deep}
Kawaguchi, K. (2016).
\newblock Deep learning without poor local minima.
\newblock In {\em Advances in neural information processing systems}, pages
  586--594.

\bibitem[Keskar et~al., 2017]{keskar2016large}
Keskar, N.~S., Mudigere, D., Nocedal, J., Smelyanskiy, M., and Tang, P. T.~P.
  (2017).
\newblock On large-batch training for deep learning: Generalization gap and
  sharp minima.
\newblock {\em International Conference on Learning Representations}.

\bibitem[Keskar and Socher, 2017]{keskar2017improving}
Keskar, N.~S. and Socher, R. (2017).
\newblock Improving generalization performance by switching from adam to sgd.
\newblock {\em arXiv preprint arXiv:1712.07628}.

\bibitem[Kingma and Ba, 2015]{kingma2014adam}
Kingma, D.~P. and Ba, J. (2015).
\newblock Adam: A method for stochastic optimization.
\newblock {\em International Conference on Learning Representations}.

\bibitem[Kleinberg et~al., 2018]{kleinberg2018alternative}
Kleinberg, R., Li, Y., and Yuan, Y. (2018).
\newblock An alternative view: When does sgd escape local minima?
\newblock {\em International Conference on Machine Learning}.

\bibitem[LeCun et~al., 2012]{lecun2012efficient}
LeCun, Y.~A., Bottou, L., Orr, G.~B., and M{\"u}ller, K.-R. (2012).
\newblock Efficient backprop.
\newblock In {\em Neural networks: Tricks of the trade}, pages 9--48. Springer.

\bibitem[Li et~al., 2018]{li2018visualizing}
Li, H., Xu, Z., Taylor, G., Studer, C., and Goldstein, T. (2018).
\newblock Visualizing the loss landscape of neural nets.
\newblock In {\em Advances in Neural Information Processing Systems}, pages
  6389--6399.

\bibitem[Loshchilov and Hutter, 2019]{loshchilov2019decoupled}
Loshchilov, I. and Hutter, F. (2019).
\newblock Decoupled weight decay regularization.
\newblock {\em International Conference on Learning Representations}.

\bibitem[Maddox et~al., 2019]{maddox2019simple}
Maddox, W.~J., Izmailov, P., Garipov, T., Vetrov, D.~P., and Wilson, A.~G.
  (2019).
\newblock A simple baseline for bayesian uncertainty in deep learning.
\newblock In {\em Advances in Neural Information Processing Systems}, pages
  13153--13164.

\bibitem[Mohri and Rostamizadeh, 2009]{mohri2009rademacher}
Mohri, M. and Rostamizadeh, A. (2009).
\newblock Rademacher complexity bounds for non-iid processes.
\newblock In {\em Advances in Neural Information Processing Systems}, pages
  1097--1104.

\bibitem[Mohri et~al., 2018]{mohri2018foundations}
Mohri, M., Rostamizadeh, A., and Talwalkar, A. (2018).
\newblock {\em Foundations of machine learning}.
\newblock MIT press.

\bibitem[Neyshabur et~al., 2017a]{neyshabur2017exploring}
Neyshabur, B., Bhojanapalli, S., McAllester, D., and Srebro, N. (2017a).
\newblock Exploring generalization in deep learning.
\newblock In {\em Advances in neural information processing systems}, pages
  5947--5956.

\bibitem[Neyshabur et~al., 2017b]{neyshabur2017pac}
Neyshabur, B., Bhojanapalli, S., and Srebro, N. (2017b).
\newblock A pac-bayesian approach to spectrally-normalized margin bounds for
  neural networks.
\newblock {\em arXiv preprint arXiv:1707.09564}.

\bibitem[Patel, 2017]{patel2017impact}
Patel, V. (2017).
\newblock The impact of local geometry and batch size on stochastic gradient
  descent for nonconvex problems.
\newblock {\em arXiv preprint arXiv:1709.04718}.

\bibitem[Poggio et~al., 2017]{poggio2017and}
Poggio, T., Mhaskar, H., Rosasco, L., Miranda, B., and Liao, Q. (2017).
\newblock Why and when can deep-but not shallow-networks avoid the curse of
  dimensionality: a review.
\newblock {\em International Journal of Automation and Computing},
  14(5):503--519.

\bibitem[Radosavovic et~al., 2020]{radosavovic2020designing}
Radosavovic, I., Kosaraju, R.~P., Girshick, R., He, K., and Doll{\'a}r, P.
  (2020).
\newblock Designing network design spaces.
\newblock In {\em Proceedings of the IEEE/CVF Conference on Computer Vision and
  Pattern Recognition}, pages 10428--10436.

\bibitem[Rangamani et~al., 2019]{rangamani2019scale}
Rangamani, A., Nguyen, N.~H., Kumar, A., Phan, D., Chin, S.~H., and Tran, T.~D.
  (2019).
\newblock A scale invariant flatness measure for deep network minima.
\newblock {\em arXiv preprint arXiv:1902.02434}.

\bibitem[Sandler et~al., 2018]{sandler2018mobilenetv2}
Sandler, M., Howard, A., Zhu, M., Zhmoginov, A., and Chen, L.-C. (2018).
\newblock Mobilenetv2: Inverted residuals and linear bottlenecks.
\newblock In {\em Proceedings of the IEEE conference on computer vision and
  pattern recognition}, pages 4510--4520.

\bibitem[Shorten and Khoshgoftaar, 2019]{shorten2019survey}
Shorten, C. and Khoshgoftaar, T.~M. (2019).
\newblock A survey on image data augmentation for deep learning.
\newblock {\em Journal of Big Data}, 6(1):60.

\bibitem[Smith, 2017]{smith2017cyclical}
Smith, L.~N. (2017).
\newblock Cyclical learning rates for training neural networks.
\newblock In {\em 2017 IEEE Winter Conference on Applications of Computer
  Vision (WACV)}, pages 464--472. IEEE.

\bibitem[Sontag, 1998]{sontag1998vc}
Sontag, E.~D. (1998).
\newblock Vc dimension of neural networks.
\newblock {\em NATO ASI Series F Computer and Systems Sciences}, 168:69--96.

\bibitem[Szegedy et~al., 2016]{szegedy2016rethinking}
Szegedy, C., Vanhoucke, V., Ioffe, S., Shlens, J., and Wojna, Z. (2016).
\newblock Rethinking the inception architecture for computer vision.
\newblock In {\em Proceedings of the IEEE conference on computer vision and
  pattern recognition}, pages 2818--2826.

\bibitem[Tan and Le, 2019]{tan2019efficientnet}
Tan, M. and Le, Q.~V. (2019).
\newblock Efficientnet: Rethinking model scaling for convolutional neural
  networks.
\newblock {\em arXiv preprint arXiv:1905.11946}.

\bibitem[Wang et~al., 2018]{wang2018identifying}
Wang, H., Keskar, N.~S., Xiong, C., and Socher, R. (2018).
\newblock Identifying generalization properties in neural networks.
\newblock {\em arXiv preprint arXiv:1809.07402}.

\bibitem[Wen et~al., 2018]{wen2018smoothout}
Wen, W., Wang, Y., Yan, F., Xu, C., Wu, C., Chen, Y., and Li, H. (2018).
\newblock Smoothout: Smoothing out sharp minima to improve generalization in
  deep learning.
\newblock {\em arXiv preprint arXiv:1805.07898}.

\bibitem[Zagoruyko and Komodakis, 2016]{zagoruyko2016wide}
Zagoruyko, S. and Komodakis, N. (2016).
\newblock Wide residual networks.
\newblock {\em arXiv preprint arXiv:1605.07146}.

\bibitem[Zhang et~al., 2019]{zhang2019cyclical}
Zhang, R., Li, C., Zhang, J., Chen, C., and Wilson, A.~G. (2019).
\newblock Cyclical stochastic gradient mcmc for bayesian deep learning.
\newblock {\em arXiv preprint arXiv:1902.03932}.

\bibitem[Zhou et~al., 2020]{zhou2020towards}
Zhou, P., Feng, J., Ma, C., Xiong, C., HOI, S., et~al. (2020).
\newblock Towards theoretically understanding why sgd generalizes better than
  adam in deep learning.
\newblock {\em arXiv preprint arXiv:2010.05627}.

\end{thebibliography}
\bibliographystyle{apalike}

\end{document}